\documentclass[final,3p,times]{elsarticle}
\usepackage{mathtools}

\usepackage{amssymb}
\usepackage{subcaption}

\usepackage{amsthm}
\usepackage{amsmath}
\usepackage{enumitem}
\usepackage{booktabs}
\usepackage{multirow}
\usepackage{lscape}
\usepackage{tabu}
\usepackage{threeparttable}
\usepackage{natbib}

\usepackage{tikz}
\usetikzlibrary{shapes.geometric, arrows}
\tikzstyle{startstop} = [rectangle, rounded corners, minimum width=3cm, minimum height=1cm,text centered, draw=black, fill=red!30]
\tikzstyle{input} = [trapezium, trapezium left angle=60,trapezium right angle=120, minimum width=1cm, minimum height=1cm, text centered, draw=black, fill=blue!5, text width=2cm]
\tikzstyle{output} = [trapezium, trapezium left angle=60,trapezium right angle=120, minimum width=1cm, minimum height=1cm, text centered, draw=black, fill=blue!20, text width=3cm]
\tikzstyle{NN} = [trapezium, minimum width=4.5cm, minimum height=1cm, text centered, draw=black, fill=orange!60]

\tikzstyle{process} = [rectangle, rounded corners, minimum width=4.5cm, minimum height=1cm, text centered, draw=black, fill=orange!30]
\tikzstyle{decision} = [diamond, minimum width=3cm, minimum height=1cm, text centered, draw=black, fill=green!5]
\tikzstyle{arrow} = [thick,->,>=stealth]
\tikzstyle{arrow2} = [dashed,->,>=stealth]

\usepackage[pagewise]{lineno}

\usepackage{setspace}

\journal{Accident Analysis and Prevention Journal}

\begin{document}

\begin{doublespacing}
\begin{frontmatter}
    
    
    
    \title{\textbf{CGAN-EB: A Non-parametric Empirical Bayes Method for Crash Hotspot Identification Using Conditional Generative Adversarial Networks: A Real-world Crash Data Study \\
    % CGAN-EB: A Deep Neural Network Based Crash Predictive Model Combined with Empirical Bayes Method for Crash Hotspot Identification
    % Application of Conditional Generative Adversarial Networks in Developing Crash Predictive Models for Road Network Safety Screening 
    }}
    
    
    \author{Mohammad Zarei $^1$}
    \author{Bruce Hellinga $^2$}
    \author{Pedram Izadpanah $^3$}

    \address{$^1$ Graduate Research Assistant, Department of Civil and Environmental Engineering, University of Waterloo, 200 University Ave., Waterloo, ON N2L3G1, Canada (corresponding author). E-mail: mzarei@uwaterloo.ca}
    \address{$^2$ Professor, Department of Civil and Environmental Engineering, University of Waterloo, 200 University Ave., Waterloo, ON N2L3G1, Canada. E-mail: bruce.hellinga@uwaterloo.ca }
    \address{$^3$ Adjunct Assistant Professor, Department of Civil and Environmental Engineering, University of Waterloo, 200 University Ave., Waterloo, ON N2L3G1, Canada. E-mail: pedram.izadpanah@uwaterloo.ca}

    \begin{abstract}
        % \begin{linenumbers}
    The empirical Bayes (EB) method based on parametric statistical models such as the negative binomial (NB) has been widely used for ranking sites in road network safety screening process. This paper is the continuation of the authors’ previous research, where a novel non-parametric EB method for modeling crash frequency data data based on Conditional Generative Adversarial Networks (CGAN) was proposed and evaluated over several simulated crash data sets. Unlike parametric approaches, there is no need for a pre-specified underlying relationship between dependent and independent variables in the proposed CGAN-EB  and they are able to model any types of distributions. The proposed methodology is now applied to a real-world data set collected for road segments from 2012 to 2017 in Washington State. The performance of CGAN-EB in terms of model fit, predictive performance and network screening outcomes is compared with the conventional approach (NB-EB) as a benchmark. The results indicate that the proposed CGAN-EB approach outperforms NB-EB in terms of prediction power and hotspot identification tests.
        
        % \end{linenumbers}
    \end{abstract}
    

\end{frontmatter}
\end{doublespacing}

% \linenumbers

\section{Introduction and motivation}
\label{section:intro}

Network screening (also known as hotspot identification) is an essential component of road safety improvement programs \cite{HSM}, in which the safety performance of sites (e.g., intersections, road segments) is evaluated followed by their ranking based on some risk measures. A common approach to estimate the crash risk at individual sites is using parametric modeling within an empirical Bayes (EB) framework that can correct for regression-to-the-mean bias \cite{hauer1997observational}. In this approach historical records are used to develop a crash predictive model that relates crash frequency as a function of variables such as road characteristics and traffic exposure. The number of crashes predicted by the model is then combined with the site's historical crash records through an EB framework \cite{hauer2002tutorial}.

Because of its ease of implementation in conjunction with the EB method and its ability to handle crash data over-dispersion, the Negative Binomial (NB) model is a commonly used parametric model in practise. Poisson-lognormal \cite{miranda2005possionlognormal}, Sichel model \cite{zou2013Sichel}, and NB-Lindley \cite{geedipally2012NB-Lindley} are some other parametric models that can account for over-dispersion and can be implemented within the EB framework. Each of these models may outperform the others in different data sets, making model selection an important step.

In addition to being technically demanding and time-consuming, selecting the optimal parametric model and constructing a functional form that best matches the given data can have a considerable impact on the outcome of network screening \cite{spf_form1,spf_form2}. This can be avoided by using a non-parametric data-driven model such as deep neural networks. The fitting and predictive performance of such models has been proven in several studies to be superior to that of parametric models \cite{dong2018NN,zeng2016NN,huang2016NN, pan2017NN, karlaftis2011statisticalvsNN, singh2020deepNN}. Due to their ability to extract underlying features relationships in the data, these computational models can reduce the work for feature selection and feature engineering, which is essential for parametric models. There is, however, no method in the literature that employs a non-parametric model, such as a deep neural network, for EB estimation and network screening.

In our previous study \cite{zarei2021Sim}, a novel EB estimation method called CGAN-EB was proposed that is based on a recent class of deep learning models called Conditional Generative Adversarial Network (CGAN) \cite{mirza2014cgan} which has been widely used in modeling complex data sets (e.g. images). The evaluation results showed that CGAN-EB can outperform the conventional method, NB-EB in terms of hotspot identification performance and EB estimate accuracy. 

The primary goal of this study is to evaluate the performance of CGAN-EB over a real-world crash data set. To this end, crash data from Highway Safety Information Systems (HSIS) collected for divided urban 4-lane freeways from 2012 to 2017 in Washington State were used to develop and compare the models.

The remainder of this paper is organized as follows. The next section reviews relevant  non-parametric models that have been used for crash data modeling and describes the CGAN model that is used in this paper. Section \ref{section:Methodology} describes the CGAN-EB model proposed for network screening along with the NB-EB model used as the benchmark model. The crash data set is described in Section \ref{section:data}. The results are presented and discussed in section \ref{section:result}. Finally, conclusions and recommendations for future works are presented in section \ref{section:conclusion}.

\section{Background}
\label{section:background}
Developing a reliable crash predictive model (also called a safety performance function (SPF)), is a complex process owing to some inherent characteristics of crash data such as excess zero values (i.e. low sample mean), temporal/spatial correlation, multicollinearity (i.e. the high degree of correlation between two or more independent variables), and over-dispersion (i.e. variance is greater than mean) to name a few \cite{lord2010crashdata}. Several statistical modeling approaches have been used to deal with these issues that are mainly based on Poison and Negative Binomial (NB) models or their variants in the form of mixture models or zero-inflated models \cite{zou2018mixNB, ye2018SNP, lord2005zeroinf}. Practically, it is recommended that the Poisson regression model is estimated as an initial model, and if over-dispersion is found, then both negative binomial and zero-inflated count models could be considered \cite{lee2002modeling}.

Besides the considerable effort required for feature engineering/selection and model selection, the traditional parametric modeling approach has strict hypotheses regarding the error terms and underlying relationship between dependent and independent variables \cite{lord2010crashdata}. In addition, they usually have difficulties dealing with outliers, missing or noisy data as well as handling multicollinearity and complex non-linearity \cite{karlaftis2011statisticalvsNN}. 

Non-parametric approaches, such as deep learning models, have been applied to different traffic safety problems to overcome these limitations. Some of the recent applications of deep learning models in crash data analysis include a crash count model with an embedded multivariate negative binomial model \cite{dong2018NN}, developing a global safety performance function \cite{pan2017NN}, real time crash predictions \cite{theofilatos2019realtime, li2020real}, pedestrian near-accident detection \cite{zhang2020pedstrian}, crash severity prediction \cite{rezapour2020severity,zheng2019severity}, and crash data augmentation \cite{islam2021vae, cai2020real}.

In terms of crash predictive models, several previous studies have compared the predictive performance of neural network models with NB models. For example, it has been shown that a simple artificial neural network (ANN) with  three-layer (i.e. input layer, hidden layer, and output layer) can provide slightly more accurate estimations for crash counts than the NB model \cite{chang2005NNvsNB}. The ANN model performed better for road segments with one or more crashes while the NB model preformed slightly better for sections with zero crash counts. In another study, a deep neural network with four layers was trained to develop a single global safety performance function for road segments \cite{pan2017NN}. The results suggest that a single deep learning model trained over multiple data sets provides a level of predictive accuracy that is at least comparable to the traditional NB model when a separate model is developed for each data set separately. A more recent study compared a six-layer neural network model with a random effect negative binomial model, a type of NB model that assumes the over-dispersion parameter is randomly distributed across groups \cite{singh2020deepNN}. The performance indicators (root mean square error, mean absolute error and correlation coefficient) confirmed that the deep neural network (DNN) model provided better performance than the parametric model. 

One criticism of DNN models is that they are perceived as black-boxes, which cannot provide the interpretable parameters as statistical models do. However, a recent study proposed a Visual Feature Importance (ViFI) method that can mitigate this limitation of DNN models \cite{pan2020VIFi}. The proposed method allows for the accurate deciphering of the model's inner workings, as well as the identification of significant features and the elimination of unnecessary features.

Although the aforementioned studies have presented different approaches to develop DNN crash predictive models, there has not been any investigation of how the EB method can be used with a DNN model. In this paper, using a powerful deep learning technique called Conditional Generative Adversarial Network (CGAN), a DNN-based model is combined with the EB method and compared with the traditional NB-EB approach in terms of model fitting,  predictive performance and network screening results. In the next section the concept of CGAN is briefly described. 

\section{Conditional Generative Adversarial Network (CGAN)}
\label{section:CGAN}
Generative Adversarial Networks (GANs) are a relatively recently developed type of deep generative models from the unsupervised machine learning field that can implicitly model any kind of data distribution and generate new samples and has achieved tremendous success in many fields (e.g., image/video synthesis/manipulation, natural language processing, classification) in recent years \cite{goodfellow2014GAN,goodfellow2016GANtutorial,creswell2018ganreview}. GANs consist of simultaneously training two deep neural networks, a generator which produces synthetic samples that mimic the characteristics (i.e. distribution) of the real (observed) data, and a discriminator which tries to distinguish between the synthetic samples coming from the generator and the real samples from the original data set. This competition between the generator and discriminator has been formalized as a min-max optimization problem and shown that at Nash equilibrium of the contest between the generator and the discriminator, the generator can capture the  distribution of the observed data (for more information refer to \cite{goodfellow2014GAN}. 

\begin{figure}[]
    \centering
    
    \begin{tikzpicture}[auto, node distance=4cm,>=latex']

        \node (G) [NN, yshift = 0cm] {\textbf{Generator}};
        \node (G_out) [output, above of = G, xshift = 0cm, yshift=-2cm, text width = 1.5cm] {\footnotesize $G(X, z)=\hat{y}$};
        
        \node (noise) [input, left of = G, yshift = 0.8cm, text width = 1cm] {\footnotesize Noise ($z$)};
        \node (X) [input, left of = G, yshift = -0.8cm, text width = 0.25cm] {\footnotesize $X$};
        
        \node (D) [NN, above of= G_out, yshift=-2cm] {\textbf{Discriminator}};
        
        \node (real) [input, above of=D, yshift=-2.5cm, text width = 1.5cm] {\footnotesize Real  data $(X, y)$};
        
        \node (D_out_real) [output, left of= D, yshift = -1.5cm, xshift = -1cm, text width = 1.5cm] {\footnotesize Real Loss: $\mathbf{L(}D(X,y), \mathbf{1})$};
        \node (D_out_fake) [output, right of= D, yshift = -1.5cm, xshift = 1cm, text width = 1.5cm] {\footnotesize Fake Loss: $\mathbf{L(}D(X,\hat{y}), \mathbf{0})$};
        
        \draw [arrow] (noise) |- (G);
        \draw [arrow] (X) |- (G);
        \draw [arrow] (G) -- (G_out);
        \draw [arrow] (G_out) -- (D);
        \draw [arrow] (real) -- (D);
        \draw [arrow] (D) -- (D_out_real);
        \draw [arrow] (D) -- (D_out_fake);
        \draw [arrow2] (D_out_fake) |- (G);
        \draw [arrow2] (D_out_fake) |- (D);
        \draw [arrow2] (D_out_real) |- (D);
 
    \end{tikzpicture}
    
    \caption{CGAN training structure ($X$ is feature vector, $y$ is the dependent variable, $z$ is a noise value from a normal distribution $N(0,1)$)}
    \label{fig:CGAN}
\end{figure}
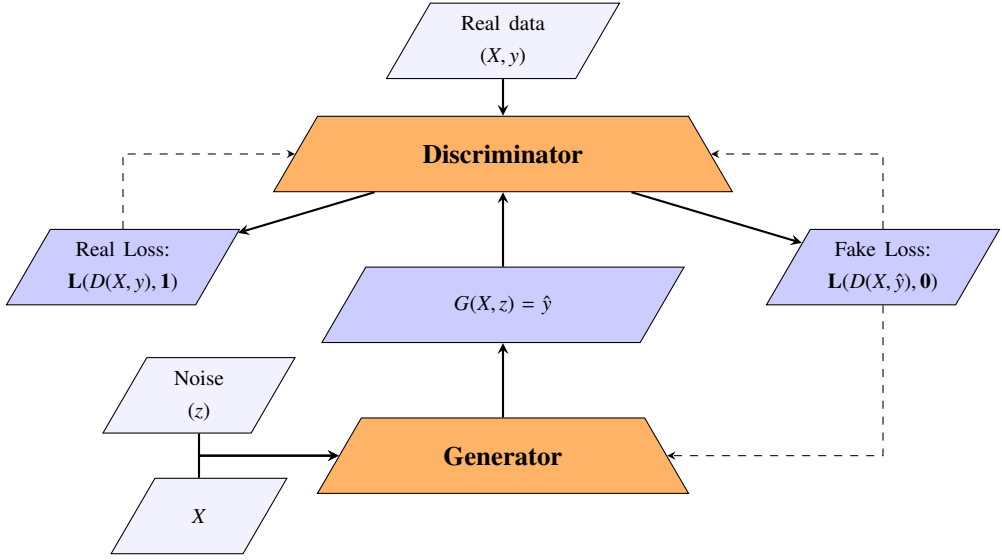

In a conditional-GAN \cite{mirza2014cgan}, referred to as CGAN, both the generator and the discriminator are conditioned on some data which could be a class label or a feature vector if we wish to use it for regression purposes. For the regression, the training steps are presented in Figure \ref{fig:CGAN}. 

The CGAN training begins by assigning two sets of random weights to both the discriminator and generator neural networks. Next, real loss value which shows the ability of the discriminator to recognize the real instances (i.e. $y$) is calculated based on $D(X, y)$, unit vector (i.e. $\mathbf{1}$) and a loss function (e.g. binary cross entropy). Note that $D(X, y) \in [0,1]$ is the output of the discriminator using $(X, y)$ as input and $G(X,z)=\hat{y}$ is the output of generator using $(X,z)$ as input. Fake loss value which shows the ability of the discriminator to recognize fake instances (i.e. $\hat{y}$) is calculated based on $D(X, \hat{y})$, zero vector (i.e. $\mathbf{0}$) and a loss function (e.g. binary cross entropy). Here $D(X,\hat{y})$ is the output of the discriminator using $(X,\hat{y})$ as input. Then, the weights of the discriminator are updated based on the objective to minimize total loss (i.e. real loss + fake loss) and the weights of the generator will be updated based on its objective to maximize fake loss. This cycle continues until the stop condition (e.g. maximum number of epochs) is met. In an ideal training condition, both fake loss and real loss converges to 0.5 which indicates that it is impossible to distinguish between input real data and synthetic data because they are samples of the same distribution \cite{mirza2014cgan}. In the context of crash prediction models, $X$, $y$ and $\hat{y}$ represent site characteristics, crash count, and the generated crash count by generator respectively.

The generator mimics the underlying distribution of real data so it can generate $m$ samples ($y_{ij}; j=1, m$) for site $i$ and this set of samples is conditioned on the feature vector $X_i$. The mean of the samples can  then be interpreted as the prediction of the model \cite{aggarwal2019cganreg}.

GAN and its variants have been rarely used in crash data analysis. The review of the literature indicates very little application of GAN and its variants to crash data analysis problems.  It has been observed that GAN is used to generate traffic data related to crashes in order to overcome data imbalance in real-time crash prediction \cite{cai2020real}. In other transportation areas, GAN have been recently used for network traffic prediction \cite{zhang2019trafficgan} and traffic flow data imputations \cite{chen2019traffic}. In this study, we propose an EB framework using CGAN, named  CGAN-EB, as an alternative to NB models for SPFs in the network screening process.

\section{CGAN-EB framework and evaluation methods}
\label{section:Methodology}

As proposed by Hauer \cite{hauer1997observational}, given $K$ as the observed number of crashes and $k$ as the expected crash count, the EB estimator of $k$ can be calculated as follows:
\begin{equation}
    \label{Eq:EB}
    E(k|K) = w \times E(k) + (1-w)\times K \approx w \times E(K) + (1-w)\times K
\end{equation}

\noindent where the weight $w$ is a function of the mean and variance of $k$ and is always a number between 0 and 1:
\begin{equation}
    \label{Eq:w}
    w = \frac{E(k)}{E(k) + VAR(k)} 
\end{equation}

\noindent In the above equations, if $k$ is gamma distributed, then the resulting $K$ follows an NB distribution (Eq. \ref{Eq:NB}). As a result, the NB-EB estimate can be derived as follows:

\begin{equation}
    \label{Eq:NB}
    f(y|\mu, \alpha) = \frac{\Gamma(y+1/\alpha)}{\Gamma(1/\alpha)\Gamma(y+1)}\left( \frac{\alpha \mu}{1+\alpha \mu}\right)\left( \frac{1}{1+\alpha \mu}\right)
\end{equation}

\begin{equation}
    \label{Eq:NB_EB}
    EB^{NB} = w \times \mu + (1-w) \times y
\end{equation}
\begin{equation}
    w = \frac{1}{1+\alpha \mu}
\end{equation}

\noindent where $y$ is the observed number of crashes per year, $\alpha$ is  dispersion parameter, and $\mu$ is the number of crashes predicted by the NB model.

\noindent Note that the mean and the variance of $y$ are $E[y] = \mu$ and $var(y) = \mu + \alpha \mu^2$ respectively. If $\alpha \rightarrow 0$, the crash variance equals the crash mean and the NB distribution converges to the Poisson distribution.

In order to derive the EB estimates using a CGAN model ($EB^{CGAN}$), we need $E(k)$ and $VAR(k)$ (see Eq. \ref{Eq:w}) which can be approximated using the samples (e.g. $m=500$ samples) taken from a trained CGAN model given the feature vector ($X$) of any given site:

\begin{equation}
    \label{Eq:CGAN mean}
    E^{CGAN}(k) \approx \frac{\sum_{j=1}^m CGAN_j(X)}{m}
\end{equation}
\begin{equation}
    \label{Eq:CGAN var}
    Var^{CGAN}(k) \approx \frac{\sum_{j=1}^m (E^{CGAN}(k)-CGAN_j(X))^2}{m-1}
\end{equation}

\noindent where $CGAN_j(X)$ is the $j$-th sample from total $m$ samples obtained from the CGAN model when provided with $X$, the feature vector of a given site, as input. Also, the weight factor $w$ in Eq. \ref{Eq:EB} when using a CGAN model can be defined as follows:

\begin{equation}
\label{Eq:CGAN w}
    w^{\small CGAN} \approx \frac{E^{CGAN}(k)}{E^{CGAN}(k)+ Var^{CGAN}(k)}
\end{equation}

Based on the above equations, $EB^{CGAN}$ can be formulated as:

\begin{equation}
    \label{Eq:EB CGAN}
    EB^{CGAN} = w^{\small CGAN} \times E^{CGAN}(k) + (1-w^{\small CGAN})\times k
\end{equation}

\subsection{Evaluation methods}
In order to compare the performance of CGAN and NB, three approaches have been employed. First, the fit performance of the models have been evaluated using three common criteria including mean absolute error (MAE), mean absolute percentage error (MAPE) and coefficient of determination ($R^2$ score). Second, the predictive performance of the models are assessed using a test set (e.g. crash data of another period). Third, in terms of network screening performance, we have utilized four tests including the Site Consistency Test (SCT), the Method Consistency Test (MCT), and Rank Difference Test (RDT) that are presented by \cite{cheng2008test}, and the Prediction Difference Test (PDT) which has been proposed in this study. All these tests are based on the assumption that, in the absence of significant changes, detected hotspots in time period $i$ should remain hazardous in the subsequent time period $i+1$. We note that there are two other tests called false identification test and Poisson mean difference test which are suitable for simulated crash data but they are not recommended when using real crash data sets \cite{lord2021highway}. 

The Site Consistency test (SCT) was designed to to measure the ability of a network screening method to consistently identify a hazardous site (i.e. hotspot) in subsequent observational periods. The higher the SCT score, the better the network screening method is. The SCT score for roadway segments is calculated using the following equation \cite{lord2021highway}:

\begin{equation}
    SCT_v = \frac{\sum_{r=1}^R C_{r,v,i+1}}{\sum_{r=1}^R L_{r,v}}
\end{equation}
where $C_{r,v,i+1}$ is the number of crashes at a site in the time period $i+1$ that is ranked $r$ as identified by method $v$, $L_{r,v}$ is the corresponding length of $r^{th}$ ranked site, and $R$ is the rank threshold that is used as cut-off in the hotspot identification.

The method consistency test (MCT) measures the number of commonly detected hotspots in time period $i$, and in the subsequent time period $i+1$. The greater the MCT score, the more reliable and consistent the method is. The MCT score can be expressed as:

\begin{equation}
    MCT_v = |\{x_{r=1}, x_{r=2}, ... , x_{r=R}\}_i \cap \{x_{r=1}, x_{r=2}, ... , x_{r=R}\}_{i+1}|
\end{equation}

where $x_{r=1}, x_{r=2}, ... , x_{r=R}$ are the hotpots that are ranked $1; 2;...; R$ respectively by the network screening method $v$.

In Rank Difference Test (RDT), the ranking of hotspots in two consecutive periods are compared. The method with a smaller RDT score is considered a superior method. This score can be calculated as follows:

\begin{equation}
    RDT_v = \sum_{r=1}^R |r-R(x_{r,v,i+1})|
\end{equation}
where $R(x_{r,v,i+1})$ is the rank of hotspot $x$ in period $i+1$ which has been ranked $r^{th}$ in period $i$ by method $v$.

Finally, the Prediction Difference Test (PDT) which is a revised version of total performance test \cite{jiang2014application}, checks how similar the EB estimate of the top ranked hotspots in period $i$ and period $i + 1$ are. The PDT score is computed as follows:

\begin{equation}
    PDT_v = \sum_{r=1}^R |EB_{x_r,i,v}-EB_{x_r,i+1,v}|
\end{equation}
where $EB_{x_r,i,k}$ is the EB estimate of the long-term mean of crashes for site $x_r$ with rank of $r$ in period $i$ by method $v$, and $EB_{x_r,i+1,v}$ is the corresponding EB estimate for period $i+1$. The method with a smaller PDT score is considered a superior method.

\section{Data description}
\label{section:data}
\begin{figure}[h]
    \centering
    \includegraphics[width=0.5\linewidth]{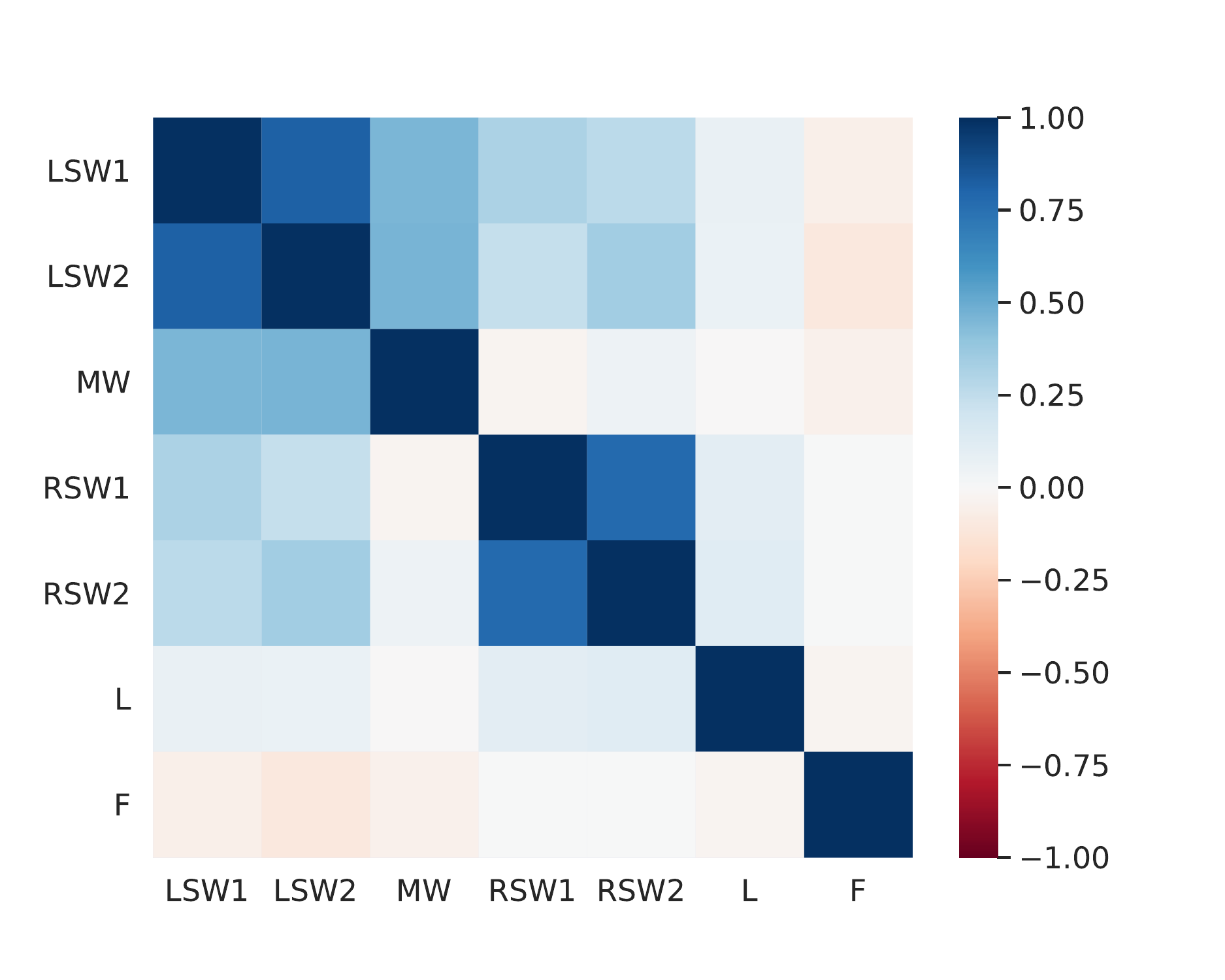}
    \caption{Variable correlation}
    \label{fig:Corr}
\end{figure}

The crash data set used in this study is from the Highway Safety Information System (HSIS) collected for divided 4-lane segments from urban freeways from 2012 to 2017 in Washington State. The data set covers a total of 3085 individual road segments with length ranging from 0.01 to 2.02 miles (0.016 to 3.25 km). 2047 segments are located in rolling terrain type and the rest (1038) are located in level type. Table \ref{Table:data1} provides a summary of statistics for seven numerical variables of the road segments in the Washington State (WA) data and the correlation of the 7 aggregated variables is displayed in Figure \ref{fig:Corr}. The blue color represents positive correlations, whereas the red tint represents negative correlations. The darker the color, the greater the correlation. It has been shown that the LSW1, LSW2, RSW1 and RSW2 are highly correlated as it was expected.

\begin{table}[ht]
\centering
\renewcommand{\arraystretch}{1.2} 
\caption{Summary of Characteristics for Individual Road Segments in the WA Data (Numerical Variables)}
    \begin{tabular}{lccc}
    \toprule
    \textbf{Numerical   Variable}                       & \textbf{Minimum} & \textbf{Maximum} & \textbf{Mean (SD) }       \\ \midrule
    Number of Crashes per year                          & 0       & 60      & 0.9 (2.5)         \\
    Left shoulder width 1   {[}LSW1  (ft){]}   & 0       & 22      & 2.1 (2.5)         \\
    Left shoulder width 2  {[}LSW2    (ft){]}  & 0       & 20      & 2.2 (2.6)           \\
    Median width {[}MW  (ft){]}                & 2       & 750     & 40.1(41.6)       \\
    Right shoulder width 1  {[}RSW1    (ft){]} & 0       & 24      & 8.6 (3.6)         \\
    Right shoulder width 2 {[}RSW2   (ft){]}   & 0       & 22      & 8.5(3.5)           \\
    Segment length {[}L (mi){]}                & 0.01    & 2.02    & 0.1 (0.1)         \\
    AADT (F)                      & 4328     & 178,149  & 46618 (27725) \\ \bottomrule
    \end{tabular}
    \begin{tablenotes}
      \centering
      \small
      \item  Note: 1 and 2 indices after LSW and RSW are related to two opposite directions of the road. Mean AADT computed over all sites and over 6 years
    \end{tablenotes}
\label{Table:data1}
\end{table}

\section{Developing models}
\label{section:Developing Models}
The data described in the previous sections were divided into two 3-year periods P1 (2012-2014) and P2 (2015-2017) in order to develop models (NB and CGAN) and compare their performances based on the described tests. The details of the model development process for the NB-EB method and the CGAN-EB method are explained in the following sections.

\subsection{NB models}
\label{section:NB models}

The NB modeling results for the WA data are provided in this section. A NB model requires the relation of crash frequency and the set of explanatory variables to be specified. We have used the following Generalized Linear Model (GLM) for this purpose:

\begin{equation}
\label{eq: functional form}
    \mu = exp(\beta_0 + \beta_L ln(L) + \beta_F ln(F) + \mathbf{\beta} \mathbf{X})
\end{equation}

\noindent where

\begin{itemize}[label={}]
    \item $\mu$ : estimated crash frequency
    \item $L$ : segment length (miles)
    \item $F$ : traffic volume (e.g. AADT)
    \item $ \mathbf{X}$ : vector of other features (UR, LSW1, LN, etc.)
    \item $\beta_L, \, \beta_F, \, \mathbf{\beta}$ : NB model coefficients
    \item $\beta_0$ : intercept
\end{itemize}

\noindent Note that, for two features $F$ (AADT) and $L$ (segment length), their natural log forms are used so that their transformations lead to the case of zero crash for zero values. This is a common functional form in crash data modeling of road segments \cite{pan2017NN, zou2013Sichel}.

The coefficients of the models have been estimated using StatsModels \cite{seabold2010statsmodels}, a Python-based module providing different classes and functions for statistical modeling and analysis. The dispersion parameter, $\alpha$ is determined using auxiliary Ordinary Least Square (OLS) regression without constant \cite{aux_OLS}. The NB model coefficients and goodness-of-fit statistics that have been estimated using maximum likelihood method  are presented in Table \ref{Table:NB results}. Note that other variables including LSW1, LSW2, RSW1, and terrain type were not statistically significant. The dispersion parameter, $\alpha$, is significantly different from zero which confirms the appropriateness of the NB model relative to the Poisson model. Consistent with expectation, the features MW (median width), F (AADT), and L (segment length) are all positively correlated with crash frequency, and increasing the width of right shoulders decreases the crash frequency.

\begin{table}[ht]
\small
\centering
\renewcommand{\arraystretch}{1.3} 
\caption{The NB Model Coefficients for the WA Data}
\begin{tabular}{lccc}
\toprule
     \textbf{Variables}     & \textbf{Coefficient ($\beta$)} & \textbf{SE}     & \textbf{p-value} \\ \midrule
Intercept & -11.8                 & 0.36  & 0.00    \\
ln(L)     & 0.902                 & 0.018 & 0.00    \\
ln(F)     & 1.33                  & 0.033 & 0.00    \\
RSW2      & -0.0627               & 0.0051 & 0.00    \\
MW        & 0.00200               & 0.00040 & 0.00    \\ \midrule
$\alpha$  & 0.836                 & 0.052  & 0.00      \\
Deviance  & 6.73E+03              & na     & na      \\
$\chi^2$  & 1.03E+04              & na     & na      \\
AIC       & 1.71E+04              & na     & na      \\
BIC       & -7.77E+04             & na     & na      \\    \bottomrule
\end{tabular}

    \begin{tablenotes}
      \small
      \item  SE = Standard Error, na = Not Available, $\alpha$: Dispersion Parameter, $\chi^2$ = Pearson’s Chi-Square Statistic
      \item AIC = Akaike Information Criteria, BIC = Bayesian Information Criterion
    \end{tablenotes}
\label{Table:NB results}
\end{table}

\subsection{CGAN models}
\label{section:CGAN models}

\begin{figure}
\small
  \centering
  \begin{subfigure}{0.48\textwidth}
    \begin{tikzpicture}[auto, node distance=4cm,>=latex']
        
        \node (concat) [process, minimum height = 0.6cm] {ConcatLayer(200)};
        \node (X_out) [NN, above of = concat, xshift=-2cm, yshift=-3cm, minimum height = 0.6cm, minimum width = 2cm] {Dense(100, ELU)};
        \node (X) [input, minimum height = 0.6cm,above of = X_out, xshift=0cm, yshift=-3cm, text width = 1cm] {$X$};
        
        \node (noise_out) [NN, above of = concat, xshift=2cm, yshift=-3cm, minimum height = 0.6cm, minimum width = 2cm] {Dense(100, ELU)};
        \node (noise) [input, minimum height = 0.6cm, above of = noise_out, xshift=0cm, yshift=-3cm, text width = 1cm] {$z$};
        
        \node (l1) [NN, below of = concat, xshift=0cm, yshift=3cm, minimum height = 0.6cm] {Dense(40, ELU)};
        \node (l2) [NN, below of = l1, xshift=0cm, yshift=3cm, minimum height = 0.6cm] {Dense(40, ELU)};
        \node (l3) [NN, below of = l2, xshift=0cm, yshift=3cm, minimum height = 0.6cm] {Dense(40, ELU)};
        \node (l4) [NN, below of = l3, xshift=0cm, yshift=3cm, minimum height = 0.6cm] {Dense(1, ReLU)};
        
         \node (Gen_out) [output, minimum height = 0.6cm,below of = l4, xshift=0cm, yshift=3cm, text width=1cm] {$\hat{y}$};

        \draw [arrow] (X) -- (X_out);
        \draw [arrow] (noise) -- (noise_out);
        \draw [arrow] (X_out) -- (concat);
        \draw [arrow] (noise_out) -- (concat);
        \draw [arrow] (concat) -- (l1);
        \draw [arrow] (l1) -- (l2);
        \draw [arrow] (l2) -- (l3);
        \draw [arrow] (l3) -- (l4);
        \draw [arrow] (l4) -- (Gen_out);

    \end{tikzpicture}
    
    \caption{Generator architecture}
    \label{fig:Generator architecture}
  \end{subfigure}
  \begin{subfigure}{0.48\textwidth}
    \begin{tikzpicture}[auto, node distance=4cm,>=latex']
        
        \node (concat) [process, minimum height = 0.6cm] {ConcatLayer(200)};
        \node (X_out) [NN, above of = concat, xshift=-2cm, yshift=-3cm, minimum height = 0.6cm, minimum width = 2cm] {Dense(100, ELU)};
        \node (X) [input, minimum height = 0.6cm, above of = X_out, xshift=0cm, yshift=-3cm, text width = 1cm] {$X$};
        
        \node (y_out) [NN, above of = concat, xshift=2cm, yshift=-3cm, minimum height = 0.6cm, minimum width = 2cm] {Dense(100, ELU)};
        \node (y) [input, minimum height = 0.6cm, above of = noise_out, xshift=0cm, yshift=-3cm, text width = 1cm] {$y$ or $\hat{y}$};
        
        \node (l1) [NN, below of = concat, xshift=0cm, yshift=3cm, minimum height = 0.6cm] {Dense(40, ELU)};
        \node (l2) [NN, below of = l1, xshift=0cm, yshift=3cm, minimum height = 0.6cm] {Dense(40, ELU)};
        \node (l3) [NN, below of = l2, xshift=0cm, yshift=3cm, minimum height = 0.6cm] {Dense(1, Sigmoid)};
        \node (Dis_out) [output, minimum height = 0.6cm,below of = l3, xshift=0cm, yshift=3cm,  text width = 4cm] {Probability of being real ($p$)};

        \draw [arrow] (X) -- (X_out);
        \draw [arrow] (y) -- (y_out);
        \draw [arrow] (X_out) -- (concat);
        \draw [arrow] (y_out) -- (concat);
        \draw [arrow] (concat) -- (l1);
        \draw [arrow] (l1) -- (l2);
        \draw [arrow] (l2) -- (l3);
        \draw [arrow] (l3) -- (Dis_out);

    \end{tikzpicture}
    
    \caption{Discriminator architecture}
    \label{fig:Discriminator architecture}
  \end{subfigure}

\caption{Network architectures. The input layer of generator includes normalized feature vector with size of 8 and a noise value ($z\sim N(0,1)$), and input layer of discriminator includes same feature vector and crash count (i.e. $y$)} 
\label{fig:architecture}
\end{figure}
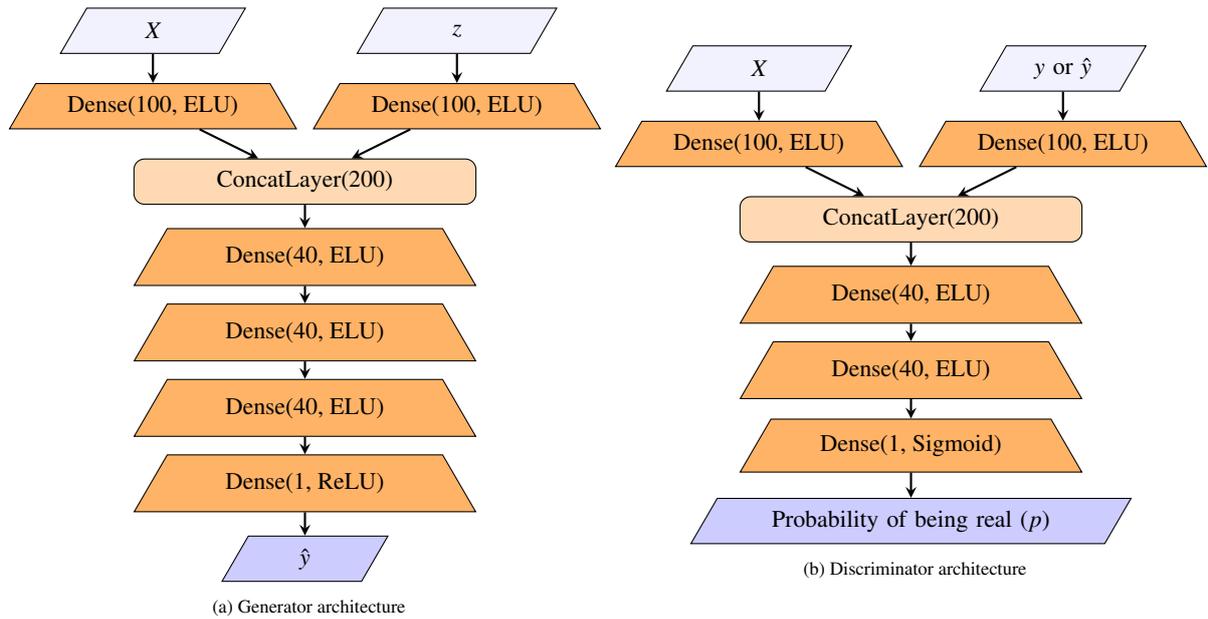

The CGAN models for this study have been developed using Keras \cite{chollet2018keras}, an open-source deep neural network library developed in Python. The architecture of the generator and discriminator are presented in Figure \ref{fig:architecture}. These architectures are designed based on suggested architectures in \cite{aggarwal2019cganreg} for using CGAN as a regression model. \textit{DenseLayer(n)} in Figure \ref{fig:architecture} is a regular deeply connected neural network layer with $n$ nodes and \textit{ConcatLayer(n)} concatenates a list of inputs. The model configuration parameters are set as follows:

\begin{itemize} [itemsep=0pt,parsep=0pt, topsep=0pt, partopsep=0pt]
    \item Activation functions: Exponential Linear Unit (ELU), Rectified Linear Unit (ReLU), and Sigmoid \cite{sharma2017activation}
    \item Optimizer: Adam \cite{pedamonti2018comparison}
    \item Number of epochs: 1000
    \item Batch size: 100
    \item Learning rate (both generator and discriminator): 0.001
    \item Learning rate decay (generator): 0.001
    \item Learning rate decay (discriminator): 0.0
\end{itemize}

\section{Results and discussion}
\label{section:result}

In this section the NB and CGAN models developed in the previous section are compared in terms of model fit, predictive performance, and network screening performance tests.

\subsection{Prediction and Fit Performance}
\label{section:Fit Performance}

In Table \ref{Table:fit results}, the NB and CGAN models are compared in terms of regression fit of each model developed on the crash data in P1. The criteria selected for this purpose include mean absolute error (MAE), mean square error (MSE) and coefficient of determination ($R^2$ score). According to all three measures, the CGAN models fit the empirical data (i.e. train data set) better than the NB models. The main reason for this is that NB models are constrained to a specified functional form in Equation \ref{eq: functional form} but there is no such constraints in CGAN models.

\begin{table}[ht]
\centering
\renewcommand{\arraystretch}{1.3} 
\caption{Regression evaluation results for CGAN and NB models for WA data}

\begin{tabular}{cccc}
\toprule
   \textbf{Criteria}      & \textbf{NB}    & \textbf{CGAN}  & \textbf{CGAN   Improvement} \\ \midrule
MAPE     & 0.56  & 0.51  & 8.9\%              \\
MAE      & 0.74  & 0.69 & 6.4\%              \\
$R^2$ score & 0.38 & 0.46 & 21\%      \\ \bottomrule     
\end{tabular}
\begin{tablenotes}
      \small
      \centering
      \item  MAE = mean absolute error, MAPE = mean absolute percentage error
    \end{tablenotes}
\label{Table:fit results}
\end{table}

The results in Table \ref{Table:fit results} evaluate the models in terms of fit to the training data (i.e. period P1) but may be impacted by over-fitting consequently the predictive performance of the NB and CGAN models are examined by comparing the accuracy of their predictions for crash counts of P2 data set as a test set. This comparison is performed to validate the models and avoid the problems of over-fitting \cite{chen2021novel}. The results are presented in Table \ref{Table: predictions perf}.

\begin{table}[ht]
\centering
\renewcommand{\arraystretch}{1.3} 
\caption{Predictive performance results for CGAN and NB models for WA data over test data set (P2)}

\begin{tabular}{cccc}
\toprule
   \textbf{Criteria}      & \textbf{NB}    & \textbf{CGAN}  & \textbf{CGAN   Improvement} \\ \midrule
MAPE     & 0.48  & 0.46  & 5.2\%              \\
MAE      & 0.91  & 0.88 & 3.6\%              \\
$R^2$ score & 0.34 & 0.39 & 15\%      \\ \bottomrule     
\end{tabular}
\begin{tablenotes}
      \small
      \centering
      \item  MAE = mean absolute error, MAPE = mean absolute percentage error
    \end{tablenotes}
\label{Table: predictions perf}
\end{table}

The results in Table \ref{Table: predictions perf} indicate that the CGAN shows better performance over the test data set (i.e. P2) which confirms the model has not over-fitted the data.

\subsection{Network Screening Performance Tests}

In order to compare the performance of NB-EB and CGAN-EB in terms of network screening results, the EB estimate for two time periods (P1 and P2) was computed for each road segment. The ranking of hotspots was based on the crash rate (i.e. the EB estimate of the crash frequency divided by the segment length).
Tables \ref{Table: 10 CGAN} and  \ref{Table: 10 NB}  present the observed crash counts over period P1 and the corresponding EB estimates by different methods for the top 10 hotspots identified by the CGAN-EB and NB-EB. Within the top 10 ranked sites, the two methods identified 9 of the same sites and these sites have a similar rank order.  As expected, the model predictions are very different from the observed crash count in some cases but the corresponding EB estimates are much closer. The reason is that both methods (CGAN-EB and NB-EB) are using crash counts in the EB estimation.

\begin{table}[]
\centering
\renewcommand{\arraystretch}{1.3} 
\caption{EB estimates and crash predictions for top 10 hotspots identified by CGAN-EB for Period P1}

\begin{tabular}{cccccccc}
\toprule
\textbf{CGAN-EB rank} & \textbf{Crash Coun}t & \textbf{L}    & \textbf{CGAN Pred.} & \textbf{CGAN-EB} & \textbf{NB Pred.} & \textbf{NB-EB} & \textbf{NB-EB rank} \\ \midrule
1              & 29          & 0.12 & 18.5            & 24.4    & 7.8           & 26.2  & 1          \\
2              & 15          & 0.06 & 4.3             & 10.5    & 3.0           & 11.6  & 2          \\
3              & 81          & 0.42 & 30.7            & 73.2    & 18.3          & 77.2  & 3          \\
4              & 23          & 0.09 & 2.5             & 14.0    & 2.2           & 15.7  & 4          \\
5              & 3           & 0.03 & 7.3             & 4.5     & 2.6           & 2.9   & 22         \\
6              & 12          & 0.06 & 3.9             & 8.3     & 2.7           & 9.1   & 5          \\
7              & 97          & 0.66 & 36.4            & 89.5    & 39.1          & 95.3  & 6          \\
8              & 72          & 0.51 & 28.1            & 65.9    & 20.5          & 69.2  & 8          \\
9              & 89          & 0.63 & 29.3            & 80.8    & 23.6          & 85.8  & 7          \\
10             & 39          & 0.3  & 25.5            & 36.7    & 13.2          & 36.9  & 9       \\ \bottomrule     
\end{tabular}
\begin{tablenotes}
      \small
      \centering
      \item  
    \end{tablenotes}
\label{Table: 10 CGAN}

\bigskip

\centering
\renewcommand{\arraystretch}{1.3} 
\caption{EB estimates and crash predictions for top 10 hotspots identified by NB-EB}

\begin{tabular}{cccccccc}
\toprule
\textbf{NB-EB rank} & \textbf{Crash Coun}t & \textbf{L}    & \textbf{CGAN Pred.} & \textbf{CGAN-EB} & \textbf{NB Pred.} & \textbf{NB-EB} & \textbf{CGAN-EB rank} \\ \midrule
1          & 29          & 0.12 & 18.5            & 24.4    & 7.8           & 26.2  & 1            \\
2          & 15          & 0.06 & 4.3             & 10.5    & 3.0           & 11.6  & 2            \\
3          & 81          & 0.42 & 30.7            & 73.2    & 18.3          & 77.2  & 3            \\
4          & 23          & 0.09 & 2.5             & 14.0    & 2.2           & 15.7  & 4            \\
5          & 12          & 0.06 & 3.9             & 8.3     & 2.7           & 9.1   & 6            \\
6          & 97          & 0.66 & 36.4            & 89.5    & 39.1          & 95.3  & 7            \\
7          & 89          & 0.63 & 29.3            & 80.8    & 23.6          & 85.8  & 9            \\
8          & 72          & 0.51 & 28.1            & 65.9    & 20.5          & 69.2  & 8            \\
9          & 39          & 0.3  & 25.5            & 36.7    & 13.2          & 36.9  & 10           \\
10         & 21          & 0.12 & 2.1             & 10.3    & 2.2           & 14.4  & 23                \\ \bottomrule     
\end{tabular}
\begin{tablenotes}
      \small
      \centering
      \item  
    \end{tablenotes}
\label{Table: 10 NB}
\end{table}

The performance of the two methods in identifying hotspots was investigated by conducting four tests that described before. The results of these tests are presented in Table \ref{Table: tests}. The values of each test for the top 2.5\%, top 5\%, top 7.5\%, and top 10\% hotspots are provided. In the last column of the table, the relative improvement of CGAN-EB over NB-EB is also presented (positive values means CGAN-EB showed better performance).

\begin{table}[ht]
\centering
\renewcommand{\arraystretch}{1.3} 
\caption{Test scores for NB-EB and CGAN-EB}

\begin{tabular}{ccccc}
\toprule
\textbf{Test}                         & \textbf{Top \% Hotspots} & \textbf{NB-EB} & \textbf{CGAN-EB} & \textbf{Improvement} \\\midrule
\multirow{4}{*}{\textbf{Site Consistency Test (SCT) score}} & 2.5\%           & 494   & 551     & 12\%             \\
                             & 5.0\%           & 390   & 415     & 6.4\%              \\
                             & 7.5\%           & 335   & 385     & 15\%           \\
                             & 10.0\%          & 308   & 332     & 7.9\%              \\ \cmidrule{2-5}
                             &                 &       & AVG     & 10\%           \\ \midrule
\multirow{4}{*}{\textbf{Method Consistency Test (MCT) score}} & 2.5\%           & 53    & 56      & 5.7\%              \\
                             & 5.0\%           & 96    & 97      & 1.0\%              \\
                             & 7.5\%           & 144   & 149     & 3.5\%              \\
                             & 10.0\%          & 194   & 201     & 3.6\%             \\\cmidrule{2-5}
                             &                 &       & AVG     & 3.4\%           \\ \midrule
\multirow{4}{*}{\textbf{Rank Difference Test (RDT) score}} & 2.5\%           & 72    & 70      & 2.8\%             \\
                             & 5.0\%           & 138   & 135     & 2.2\%              \\
                             & 7.5\%           & 176   & 170     & 3.4\%             \\
                             & 10.0\%          & 211   & 194     & 8.1\%             \\\cmidrule{2-5}
                             &                 &       & AVG     & 4.1\%           \\ \midrule
\multirow{4}{*}{\textbf{Prediction Difference Test (PDT) score}} & 2.5\%           & 6.2   & 3.6     & 41\%             \\
                             & 5.0\%           & 4.3   & 2.9     & 33\%             \\
                             & 7.5\%           & 3.5   & 2.1     & 39\%             \\
                             & 10.0\%          & 3.1   & 1.9     & 38\%      \\\cmidrule{2-5}
                             &                 &       & AVG     & 38\%           \\ \bottomrule     
\end{tabular}
\begin{tablenotes}
      \small
      \centering
      \item  
    \end{tablenotes}
\label{Table: tests}
\end{table}

The SCT scores in Table \ref{Table: tests} indicates that CGAN-EB outperformed NB-EB in identifying the top 2.5\%, 5\%, 7.5\%, and 10\% of hotspots with the highest crash frequency in P2 and improved the scores about 10\% on average. The MCT scores show a similar trend to the SCT scores and indicate improvement over all four cases with the average of 3.4\%. This means that the proposed CGAN-EB approach provides greater consistency than  NB-EB in terms of the hotspots identified in the two periods (P1 and P2). In terms of the RDT test, CGAN-EB had better performance is all hotspot levels with the average of 4.1\% improvement. On average across all four hotspot thresholds, the proposed CGAN-EB was marginally better than NB-EB.  Finally, on the basis of PDT scores, the estimates by CGAN-EB are much more consistent over the two periods than by NB-EB. All of these four tests examine the consistency of the network screening outcomes across two consecutive time periods (P1 and P2).  The superior performance of the proposed CGAN-EB method over the NB-EB suggests that it has better temporal transferability than NB-EB, but conclusive statements about temporal and spatial transferability needs further investigation in future studies.

\section{Conclusions}
\label{section:conclusion}

This paper has evaluated the performance of CGAN-EB \cite{zarei2021Sim} a novel EB estimation method based on CGAN models (CGAN-EB) - a powerful deep generative model - and compared its performance with the commonly used NB-EB model. Both approaches are used to model a crash data set collected for road segments from 2012 to 2017 in Washington State, and are compared in terms of model fit, predictive performance and network screening results. The results show that the CGAN model has better ability to fit the crash data, and provide more accurate predictions over same test data set. Moreover, four tests have been conducted to evaluate the network screening performance of CGAN-EB, and the average score across the four different thresholds of hotspots for all four tests indicate that CGAN-EB works better than NB-EB particularly  in terms of consistency of suggested hotspots and EB estimations. All of this evidence indicates that the proposed CGAN-EB method is a powerful crash modelling approach for network screening with performance that is equal to or better than the conventional NB-EB approach.  

Despite these very promising results, there remain a number of additional questions regarding the proposed method in this study that need to be investigated in the future. 
First, in the current study we were not able to make conclusions about which method (i.e. CGAN-EB or NB-EB) can more accurately identify hotspots or estimate the true long term crash rate because truth is not known.  It is only possible to carry out this type of evaluation using a simulated environment \cite{zou2015simulated} in which we know the true safety state of each site.  Furthermore, carrying out a simulation evaluation also provides an opportunity to examine the sensitivity of the CGAN-EB performance benefits (vs conventional methods) to key attributes such as the number of observations, the nature of the crash data (e.g. mean and dispersion), etc.   
Second, additional studies are required to investigate the best CGAN network configuration (i.e. architecture, size) and the sensitivity of model performance to these configurations. 
Third, transferability (both locally and temporally) of crash prediction models is highly desirable. Some previous work suggests that DNN models are more transferable than parametric models (such as the NB model). It is necessary to investigate the spatial and temporal transferability of the proposed CGAN-EB model

% \section{Acknowledgements}
% \label{section:acknowledgements}
% The authors gratefully acknowledge ..... to access the site characteristic data, traffic volume data, and crash data utilized in this study.

% \section{Data Availability Statement}
% \label{section:Data Availability Statement}

% All data, models, or code that support the findings of this study are available from the corresponding author upon reasonable request.



%% References

\bibliographystyle{Other/model1-num-names.bst}
\bibliography{main.bib}






% \end{thebibliography}

\end{document}